# Lightning Does Not Strike Twice:
# Robust MDPs with Coupled Uncertainty

**Shie Mannor**  SHIE@EE.TECHNION.AC.IL
Department of Electrical Engineering, Technion, Israel

**Ofir Mebel**  OFIRMEBEL@GMAIL.COM
Department of Electrical Engineering, Technion, Israel

**Huan Xu**  MPEXUH@NUS.EDU.SG
Department of Mechanical Engineering, National University of Singapore, Singapore

## Abstract

We consider Markov decision processes under parameter uncertainty. Previous studies all restrict to the case that uncertainties among different states are uncoupled, which leads to conservative solutions. In contrast, we introduce an intuitive concept, termed "Lightning Does not Strike Twice," to model coupled uncertain parameters. Specifically, we require that the system can deviate from its nominal parameters only a bounded number of times. We give probabilistic guarantees indicating that this model represents real life situations and devise tractable algorithms for computing optimal control policies.

## 1. Introduction

Markov decision processes (MDPs) are widely used tools to model sequential decision making in stochastic dynamic environments (e.g., Puterman, 1994; Sutton & Barto, 1998). Typically, the parameters of these models (reward and transition probability) are estimated from finite and sometimes noisy data, so they often deviate from their true value. Such deviation, termed "parameter uncertainty," can cause the performance of the "optimal" policies to degrade significantly, as demonstrated in Mannor et al. (2007).

Many efforts have been made to alleviate the effect of parameter uncertainty in MDPs (e.g., Nilim & El Ghaoui, 2005; Iyengar, 2005; Givan et al., 2000; Ep-stein & Schneider, 2007; Bagnell et al., 2001). Inspired by the so-called "robust optimization" framework (Ben-Tal & Nemirovski, 1998; Bertsimas & Sim, 2004; Ben-Tal et al., 2009), the common approach regards the MDP's uncertain parameters $r$ and $p$ as fixed but unknown elements of a known set $\mathcal{U}$, often termed as the *uncertainty set*, and ranks solutions based on their performance under (respective) worst parameter realization.

Previous study in robust MDPs typically assumes that there exists *no coupling* between uncertain parameters of different states. That is, $\mathcal{U} = \prod_{s \in \mathcal{S}} \mathcal{U}_s$, where $\mathcal{U}_s$ is the uncertainty set for $(p_s, r_s)$, the transition probabilities and rewards for each state. Such an assumption is restrictive, and can often lead to conservative solutions as it essentially assumes that *all* parameters take worst possible realization *simultaneously* – an event that rarely happens in practice.

In this paper we propose and analyze a robust MDP approach that allows *coupled* uncertainty to mitigate conservativeness of the uncoupled uncertainty model. Specifically, we consider a setup which we term LDST after famous proverb "Lightning Does not Strike Twice:" while each parameter may be any element of the uncertainty set, the total number of states whose parameters deviate from their nominal value is bounded. The motivation is intuitive: suppose the parameter of each state deviates with a small probability, and all states are independent, then the total number of states with deviated parameters will be small. Hence, protection against the scenario that *only a limited number of states* deviate from their nominal parameter is sufficient to ensure that the solution is very likely to be robust to the real parameters.

The LDST approach is conceptually intuitive and sim-ple, yet with added modeling power. Indeed, it is backed by probabilistic guarantees that hold under reasonable conditions. While the LDST approach provides a flexible way to control the conservativeness of robust MDPs, it remains computationally friendly: it is tractable even when both reward parameters and transition probabilities are uncertain, for one of the two setups which we will consider. For the other one that is computationally more challenging, it is tractable in the special case where only the reward parameters are uncertain.

One advantage of the LDST approach is that it requires no distribution information. This is in sharp contrast to some other variants of MDP methods also aiming to mitigate conservativeness, including Bayesian reinforcement learning (Strens, 2000; Poupart, 2010), chance constrained MDP (Delage & Mannor, 2010), and distributionally robust MDP (Xu & Mannor, 2010), all assuming a-priori information on the distribution of the system parameters.

In Section 2 we give background on MDPs and define the LDST formulation. In Section 3 we define the non-adaptive model and prove that it is computationally hard in the general case, but tractable if only the rewards are subject to deviations. In Section 4 we define the adaptive model and give tractable algorithms for solving both finite and infinite-horizon problems. In Section 5 we define the concept of fractional deviation and prove that the adaptive model is solvable also when fractional deviations are considered. Simulation of the adaptive model is demonstrated in Section 6 and a conclusion is given in Section 7.

## 2. Preliminaries

In this section we briefly explain our notations, some preliminaries about MDP, and the setups that we investigate. Following standard notations from Puterman (1994), we define a Markov decision process as a 6-tuple $< T, \gamma, \mathcal{S}, \mathcal{A}, p, r >$: here $T$, possibly infinite, is the decision horizon; $\gamma \in (0, 1]$ is the discount factor, and can be 1 only when $T < \infty$. The state space $\mathcal{S}$ and the action space $\mathcal{A}$ are both finite. In state $s \in \mathcal{S}$, the decision maker can pick an action $a \in \mathcal{A}$, and obtains a reward of $r(s, a)$, and the next state will be $s' \in S$ with a probability $p(s'|s, a)$. We use a subscript $s$ to denote the parameter for state $s$. For example, $r_s$ is the reward vector for different actions in state $s$. We assume that the initial state is distributed according to $\alpha(\cdot)$, i.e., the initial state is $s$ with probability $\alpha(s)$. The sets of all history-dependent randomized strategies, all Markovian randomized strategies, and all Markovian deterministic strategies are denoted by $\Pi^{HR}$, $\Pi^{MR}$ and $\Pi^{MD}$, respectively. The decision goal is to maximize the performance $X(\pi, p, r) \triangleq \mathbb{E}_\pi\{\sum_{t=1}^{T} \gamma^{t-1} r(s_t, a_t)\}$, i.e., the expected accumulative reward, under parameter $p$ and $r$.

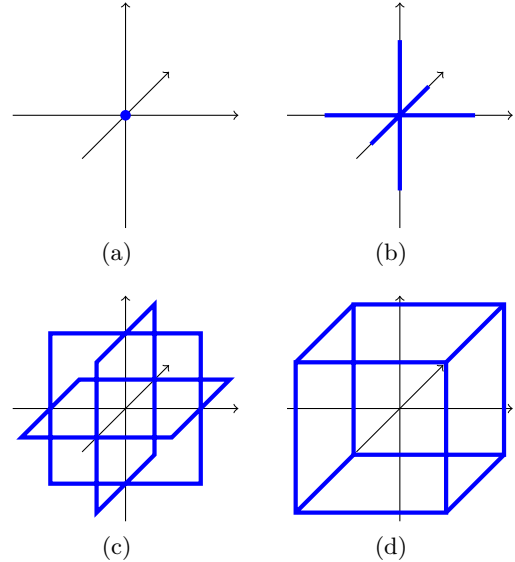

Figure 1. Illustration of effective uncertainty sets for different values of $D$. The origin symbolizes the nominal parameters $p^0$ and $r^0$: (a) $D = 0$ (no uncertainty); (b) $D = 1$; (c) $D = 2$; and, (d) $D = 3$ (uncoupled uncertainty).

An uncertain MDP (uMDP) is defined as a tuple $< T, \gamma, \mathcal{S}, \mathcal{A}, p^0, r^0, \mathcal{U}, D >$. In contrast to standard MDP, the true parameters $p^{\text{true}}$ and $r^{\text{true}}$ are unknown. Instead, for each state $s$, we know $p_s^0$ and $r_s^0$ termed *nominal parameters* hereafter, which are essentially point estimations of the parameters $p_s$ and $r_s$, as well as an uncertainty set $\mathcal{U}_s$ such that $(p_s, r_s) \in \mathcal{U}_s$. As standard in robust optimization and robust MDPs, we assume that the nominal parameters $p_s^0$ and $r_s^0$ belong to the uncertainty set $\mathcal{U}_s$, and the uncertainty sets $\{\mathcal{U}_s\}_{s \in \mathcal{S}}$ are convex. As discussed above, previous work in robust MDP all focused on the case that parameter realization is uncoupled, i.e., $(p, r)$ can take any value in $\mathcal{U} \triangleq \prod_s \mathcal{U}_s$. In this paper we add the possibility of coupling between uncertainties to avoid being overly conservative. Following the proverb "lightning does not strike twice," we consider uncertainty models that do not allow disaster to happen too frequently. In particular, besides the constraint that the parameters belong to respective uncertainty sets, we further require that the number of states whose parameters deviate from their nominal values is bounded by the given threshold $D$. An illustration of different effective uncertainty sets the model can produce is given in Figure 1 for three states, where it can be seen that the model spans both the model oblivious to uncertainty, and the conservative model of uncoupled uncertainty. We will later demonstrate that the values of $D$ in-between these extremes are both advantageous, and tractable.

Intuitively, LDST can be thought of as a two-player zero-sum game, where the decision maker chooses a policy, and an adversarial Nature picks an admissible parameter realization attempting to minimize the reward of the decision maker. Two models are possible, leading to two forms that we will investigate in the sequel. In the first model, termed *non-adaptive model*, depending on the control policy, Nature chooses the parameter realization of all states once and fixed thereafter, thus leading to a single stage game. In the second model, both players adapt their actions to the trajectory of the MDP, leading to a sequential game. To illustrate the difference, consider the following example: suppose at state $s_0$, taking an action $a$ will lead the system to either state $s_1$ or state $s_2$, both with positive probabilities, and that only the parameter of one state is allowed to deviate. In the *adaptive uncertainty model* the following deviation is allowed: deviate $s_1$ if the system reaches $s_1$, and deviate $s_2$ if the system reaches $s_2$. In contrast, in the *non-adaptive model* the state whose parameters deviate must be fixed, and can not depend on whether the system reaches $s_1$ or $s_2$.

## 3. Non-Adaptive Uncertainty Model

In this section, we focus on the non-adaptive case. As mentioned above, the uncertain parameters, while unknown, are fixed a priori and do not depend on the trajectory of the MDP. Similarly, the decision maker has to fix a strategy and can not adapt to the realization of the parameters. Equivalently, this means that the decision maker does not observe whether or when the parameters deviate.

Our goal is to find a strategy that performs best under the worst admissible parameter realization. That is, to solve the following problem:

$$\max_{\pi \in \Pi^{HR}} \min_{(p,r) \in \mathcal{U}_D} X(\pi, p, r) \qquad (1)$$

where: $\mathcal{U}_D \triangleq \left\{ (p,r) \in \mathcal{U} \,\Big|\, \sum_{s \in \mathcal{S}} \mathbf{1}_{(p_s, r_s) \neq (p_s^0, r_s^0)} \leq D \right\}.$

To put it in words, the set of admissible parameters $\mathcal{U}_D$ is defined as follows: the parameters of no more than $D$ states can deviate from their nominal values, and the deviated parameters belong to their respective uncertainty set $\mathcal{U}_s$.

The main motivation of this formulation is that often in practice, the correlation between the parameters of different states is weak, or even completely independent. In such a case, protection against the scenario that *only a limited number of states* deviate from their nominal parameter is sufficient to ensure that the solution is very likely to be robust to the real parameters. To illustrate this intuition, we have the following theorem which considers a generative deviation model where the parameters of different states are independent. The proof is deferred to Appendix A.

**Theorem 1.** *Suppose the parameters of different states deviate independently, with a probability $\alpha_s$ for $s \in \mathcal{S}$. Then with probability at least $1 - \delta$ we have $(p, r) \in \mathcal{U}_{D'}$, where $D' \triangleq \sum_{s \in \mathcal{S}} \alpha_s + \frac{1}{3} \log(1/\delta) \left( 1 + \sqrt{1 + 18 \frac{\sum_{s \in \mathcal{S}} \alpha_s}{\log(1/\delta)}} \right).$*

### 3.1. Computational Complexity

In general, the non-adaptive case – Problem 1 – is computationally hard. To make this statement formal, we consider the following "yes/no" decision problem.

**Decision Problem.** *Given a Markov decision process with $|\mathcal{S}| = n$ and $|\mathcal{A}| = m$, $D \in [0:n]$, $\mathcal{U}_s$ for all $s \in \mathcal{S}$ and $\beta \in \mathbb{R}$. Does there exist $\pi \in \Pi^{HR}$ such that*

$$\min_{(p,r) \in \mathcal{U}_D} X(\pi, p, r) \geq \beta ? \qquad (2)$$

We denote the decision problem by $\mathcal{L}(\Pi^{HR})$. Similarly we can define $\mathcal{L}(\Pi^{MR})$ and $\mathcal{L}(\Pi^{MD})$. We next show that answering the decision problem is hard even for very simple uncertainty set $\mathcal{U}$. This immediately implies that *finding* a strategy $\pi$ that satisfies (2) is computationally difficult. The proof, deferred to Appendix B, is based on reduction from *Vertex Cover Problem*.

**Theorem 2.** *Suppose for any $s \in S$, $\mathcal{U}_s$ is a line segment, i.e., the convex combination of two points. Then, the problems $\mathcal{L}(\Pi^{HR})$, $\mathcal{L}(\Pi^{MR})$ and $\mathcal{L}(\Pi^{MD})$ are NP-hard.*

### 3.2. Reward-Uncertainty Case

While in general Problem 1 can be intractable, we show that if only the reward parameters are subject to uncertainty then the problem is solvable in polynomial time. We first define the concept of tractable uncertainty set:

**Definition 1.** *A set $\mathcal{U}$ is called* tractable*, if in polynomial time, the following can be solved for any $c$:*

*Minimize: $c^\top x$;   s.t. $x \in \mathcal{U}.$*

Using this formulation we state the main result of this section. The proof is given in Appendix C in the supplementary material.

**Theorem 3.** *Suppose that only the reward is subject to uncertainty. If for all $s \in \mathcal{S}$, $\mathcal{U}_s$ is a tractable uncertainty set, then Problem 1 can be solved in polynomial time.*

When uncertainty sets are polytopes or ellipsoids as often the case in practice, we can convert Problem 1 into easier convex optimization problems such as LP or quadratic programming, for which large scale problems can be solved using standard solvers in reasonable time. See Appendix D and E for detail.

## 4. Adaptive Uncertainty Model

This section is devoted to the *adaptive case*, i.e., the parameter realization – in particular the choice of deviating states – depends on the history trajectory of the MDP. Moreover, the decision maker observes parameter deviation *retrospectively*. That is, in each decision stage the decision maker chooses an action, after which the true parameters are realized and observed by the decision maker. Hence the decision maker is aware of the occurrence of parameter deviation, and his strategy is allowed to depend on such information.

In practice, there are two cases where the agent can confidently determines a deviation. The first case is where the parameter deviation is accompanied by external signals (e.g., lighting). For example, consider a problem of determining the route for a helicopter, the deviation is when there is a storm, and it is reasonable to assume that the agent can observe the storm if it really happens. A similar example is portfolio optimization where the parameter deviation is due to market crash or company bankrupcy, which certainly can be observed. The second case, where there is no external signal, is the case where the support of the nominal parameter and deviation barely intersects (in the probabilistic sense). Take the simulation in Section 6 as an example, although under the nominal parameter, it is possible that the number of customers come is large, the probability is very small. Recurrence of such events is more unlikely. Hence, the agent can count each of these events as one parameter deviation, and allow the Nature one or two extra chances of parameter deviation. Solving the resulting adaptive case would yield a solution that with overwhelming probabilities is robust to fixed number of parameter deviation, yet not overly conservative. Notice that a special example of the second case is when the supports are disjoint.

### 4.1. Finite-horizon case

Mathematically, the finite-horizon adaptive case can be formulated as the following decision problem [1],

$$\max_{a_1 \in \mathcal{A}} \min_{(p_1, r_1) \in \mathcal{U}_{s_1}} \max_{a_2 \in \mathcal{A}} \min_{(p_2, r_2) \in \mathcal{U}_{s_2}} \cdots \quad (3)$$

$$\max_{a_T \in \mathcal{A}} \min_{(p_T, r_T) \in \mathcal{U}_{s_T}} \mathbb{E}\left\{\sum_{t=1}^{T} r_t(s_t, a_t)\right\};$$

$$\text{Subject to:} \quad \sum_{t=1}^{T} \mathbf{1}_{(p_t, r_t) \neq (p_t^0, r_t^0)} \leq D.$$

Here, $s_t$ is the randomized state at decision epoch $t$. Hence, for all $s, s' \in \mathcal{S}$ we have $\Pr(s_{t+1} = s'|s_t = s, a_t) = p_t(s'|s, a_t)$. We remark that in the adaptive case we require that the number of *decision stages* (as opposed to *states*) where the parameter deviates is bounded by $D$. Hence, the parameters of a state are allowed to take different values for multiple visits. Similarly to Theorem 1, Formulation (3) is justified by the following probabilistic guarantee. The proof is almost identical to that of Theorem 1, and omitted.

**Theorem 4.** *Suppose the parameters of different stages deviate independently, with a probability $\alpha_t$ for $t = 1, \cdots, T$. Then with probability at least $1 - \delta$ we have that the total number of deviations $\sum_{t=1}^{T} \mathbf{1}_{(p_t, r_t) \neq (p_t^0, r_t^0)}$ is upper bounded by $\sum_{t=1}^{T} \alpha_t + \frac{1}{3} \log(1/\delta) \left(1 + \sqrt{1 + 18 \frac{\sum_{t=1}^{T} \alpha_t}{\log(1/\delta)}}\right)$.*

The next theorem, which is the main result of this subsection, asserts that in sharp contrast to the non-adaptive case, the adaptive-case can be solved via backward induction.

**Theorem 5.** *Let $v_{T+1}(s, d) = 0$ for all $s \in \mathcal{S}$ and $d \in [0 : D]$, and define for $t = 1, \cdots, T$,*

$$v_t(s, d) \triangleq$$
$$\begin{cases} \max_{a \in \mathcal{A}} \left\{ \min \left[ q_t(s, d, a, p_s^0, r_s^0), \right. \right. \\ \left. \left. \min_{(p,r) \in \mathcal{U}_s} q_t(s, d-1, a, p, r) \right] \right\}, & \text{if } d \geq 1; \\ \max_{a \in \mathcal{A}} q_t(s, d, a, p_s^0, r_s^0), & \text{if } d = 0; \end{cases}$$

$$a_t^*(s, d) \triangleq$$
$$\begin{cases} \arg\max_{a \in \mathcal{A}} \left\{ \min \left[ q_t(s, d, a, p_s^0, r_s^0), \right. \right. \\ \left. \left. \min_{(p,r) \in \mathcal{U}_s} q_t(s, d-1, a, p, r) \right] \right\}, & \text{if } d \geq 1; \\ \arg\max_{a \in \mathcal{A}} q_t(s, d, a, p_s^0, r_s^0), & \text{if } d = 0; \end{cases}$$

---

[1] In a slight abuse of notation we write $(p_t, r_t)$ instead of $(p_{s_t}, r_{s_t})$.

where $q_t(s, d, a, p, r) \triangleq \sum_{s'} p(s'|s,a) v_{t+1}(s', d) + r(s,a)$. Then the optimal strategy to Problem 3 is to take $a_t^*(s, d)$ at stage $t$, state $s$, where stage-parameters are allowed to deviate at most $d$ times from this stage on.

Before proving the theorem, we first explain our intuition: we can re-formulate Problem 3 as a sequential game of perfect information on an augmented state-space. More precisely, Problem 3 can be regarded as a sequential game, where there are two players, namely the *decision maker* who attempts to maximize the total expected reward by choosing actions sequentially, and *Nature* who aims to minimize the total expected reward by choosing parameter-realization. Thus, we can expand the decision horizon into $2T$ where the decision maker makes move in each odd decision stage, and Nature makes move in each even decision stage. Furthermore, both the decision maker and Nature are aware of the number of "deviated visits" so far, which affect their optimal strategies. Hence, we need to augment the state-space with this information. Thus, we construct the following game.

Consider the following augmented state space, where we have two types of states: the *decision maker* states and the *Nature* states. The set of decision maker states is given by
$$\overline{\mathcal{S}}_D = \mathcal{S} \times [0:D],$$
and the set of Nature states is given by
$$\overline{\mathcal{S}}_N = \mathcal{S} \times [0:D] \times \mathcal{A}.$$
The entire (augmented) state space is thus
$$\overline{\mathcal{S}} = \overline{\mathcal{S}}_D \bigcup \overline{\mathcal{S}}_N.$$

A zero-sum stochastic game with time horizon $2T$ between two players on $\overline{\mathcal{S}}$ is constructed as follows. In each decision epoch $2t-1$ for $t = 1, \cdots, T$, the system state belongs to $\overline{\mathcal{S}}_D$ and only the decision maker can choose an action $a \in A$. Let the state be $(s, d)$ where $s \in S$ and $d \in [0:D]$, and suppose the decision maker takes action $a \in \mathcal{A}$, then the next state will be $(s, d, a)$, with both players get zero-reward.

In each decision epoch $2t$ for $t = 1, \cdots, T$, the system state belongs to $\overline{\mathcal{S}}_N$ and only Nature can choose an action. The action set for Nature is the following: if $d \geq 1$, Nature can either pick the nominal parameter $(p_s^0, r_s^0)$, in which case the next state will be $(s', d)$ where $s'$ is a random variable following the probability distribution $p^0(s'|s,a)$, and the decision maker gets a reward $r^0(s, a)$, which Nature simultaneously loses; alternatively, Nature can pick a parameter $(p', r') \in \mathcal{U}_s$, in which case the next state will be $(s', d-1)$ where $s'$ is a random variable following the probability distribution $p'(s'|s, a)$, and the decision maker gets a reward $r'(s, a)$, which Nature, again, simultaneously loses. If $d = 0$, Nature can only pick the nominal parameter $(p_s^0, r_s^0)$. An example of this construction is given graphically in Figure 2. It is easy to see that solving Problem 3 is equivalent to solving the aforementioned stochastic game. This leads to the proof of Theorem 5.

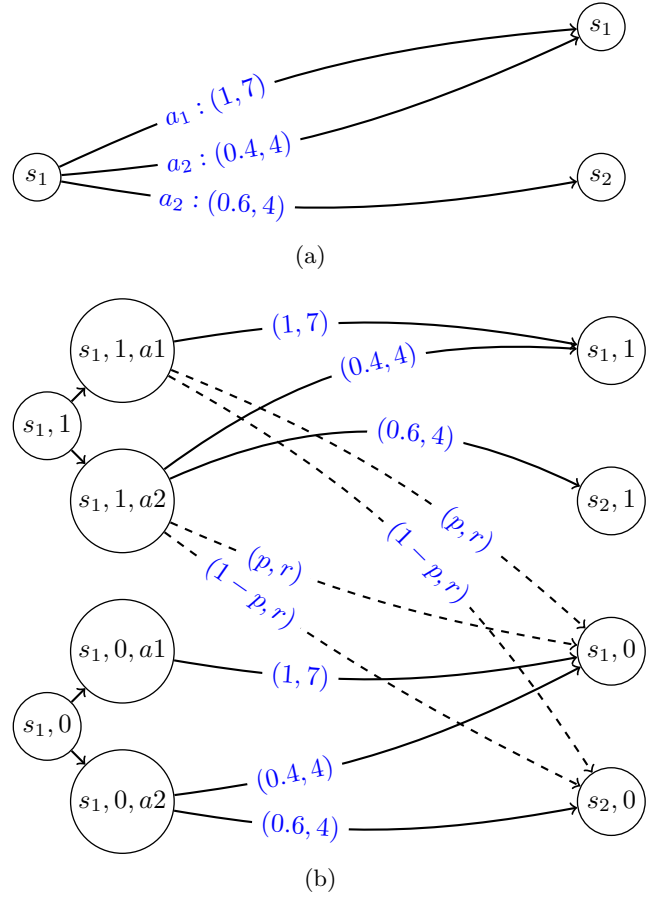

*Figure 2.* Illustration of the two-player stochastic game construction for Theorem 5, $T = 1$. (a) The nominal MDP, (b) The two-player stochastic game. Dashed lines are moves in which Nature chooses to deviate system's parameters, and $(p, r) \in \mathcal{U}_{s_1}$ is in Nature's control.

*Proof of Theorem 5.* We prove the theorem by investigating the aforementioned stochastic game, and showing that a Nash equilibrium exists for the stochastic game, where the decision maker's move is $a_t^*(s, d)$. Since the solution to the stochastic game and the solution to Problem 3 coincide (Filar & Vrieze, 1996), the theorem thus follows.

We now construct the Nash equilibrium. Let the decision maker take the strategy $a_t^*(s,d)$ at state $(s,d) \in \overline{\mathcal{S}}_D$ in decision epoch $2t-1$. Let Nature take the following action at state $(s,d,a) \in \overline{\mathcal{S}}_N$ in decision epoch $2t$: do not deviate the parameters (i.e, pick parameter $(p_s^0, r_s^0)$) if either $d=0$, or $q_t(s,d,a,p_s^0,r_s^0) \leq \min_{(p,r) \in \mathcal{U}_s} q_t(s,d-1,a,p,r)$; otherwise, pick parameters $\arg\min_{(p,r) \in \mathcal{U}_s} q_t(s,d-1,a,p,r)$.

Observe that if Nature fixes this strategy, the stochastic game reduces to a standard MDP for the decision maker to maximize his total reward, in which case his optimal strategy is take $a_t^*(s,d)$ at each $(s,d) \in \overline{\mathcal{S}}_D$ at the decision epoch $2t-1$. On the other hand, if the decision maker fixes his strategy, then for Nature to minimize the total-reward for the decision maker, it is easy to see that the aforementioned strategy is optimal. Thus, the aforementioned pair of strategies is a Nash equilibrium, which implies the theorem. $\square$

Algorithm 1 illustrates how Theorem 5 can be used to compute optimal policy for the non-adaptive model in the LDST setup. Its time-complexity is $O\left(TD|\mathcal{S}||\mathcal{A}|(|\mathcal{S}| + M)\right)$ where $M$ is defined as the maximal computational effort in performing a single minimization of $q_t(s,d,a,p,r)$ in $(p,r) \in \mathcal{U}_s$ for any $s \in \mathcal{S}, t \in [1:T]$.

### 4.2. Discounted-Reward Infinite Horizon Case

We extend the formulation of the adaptive case into the infinite horizon case, with discounted total reward criterion. Depending on whether the number of parameter deviations is also discounted, we formulate and investigate the following two setups:

**Setup A – Non-Discounted Deviates:**

$$\max_{a_1 \in \mathcal{A}} \min_{(p_1,r_1) \in \mathcal{U}_{s_1}} \max_{a_2 \in \mathcal{A}} \min_{(p_2,r_2) \in \mathcal{U}_{s_2}} \cdots \quad (4)$$

$$\max_{a_t \in \mathcal{A}} \min_{(p_t,r_t) \in \mathcal{U}_{s_t}} \cdots \mathbb{E}\left\{\sum_{t=1}^{\infty} \gamma^{t-1} r_t(s_t, a_t)\right\};$$

$$\text{Subject to:} \quad \sum_{t=1}^{\infty} \mathbf{1}_{(p_t,r_t) \neq (p_t^0, r_t^0)} \leq D.$$

**Setup B – Discounted Deviates:**

$$\max_{a_1 \in \mathcal{A}} \min_{(p_1,r_1) \in \mathcal{U}_{s_1}} \max_{a_2 \in \mathcal{A}} \min_{(p_2,r_2) \in \mathcal{U}_{s_2}} \cdots \quad (5)$$

$$\max_{a_t \in \mathcal{A}} \min_{(p_t,r_t) \in \mathcal{U}_{s_t}} \cdots \mathbb{E}\left\{\sum_{t=1}^{\infty} \gamma^{t-1} r_t(s_t, a_t)\right\};$$

$$\text{Subject to:} \quad \sum_{t=1}^{\infty} \beta^{t-1} \mathbf{1}_{(p_t,r_t) \neq (p_t^0, r_t^0)} \leq D.$$

**Algorithm 1** Backward Induction to solve the *adaptive case*

**Input:** uMDP $<T, \mathcal{S}, \mathcal{A}, p^0, r^0, \mathcal{U}, D>$
Initialize $V_{T+1}(s,d) = 0 \quad \forall s \in \mathcal{S}, d \in [0:D]$.
**for** $t = T$ **downto** $1$ **do**
  **for** $d = 0$ to $D$ **do**
    **for** $s \in \mathcal{S}$ **do**
      Initialize $bestAction = \phi$
      Initialize $bestActionValue = -\infty$
      **for** $a \in \mathcal{A}$ **do**
        {Nominal value}
        Initialize $actionValue = q_t(s,d,a,p_s^0,r_s^0)$
        **if** $d \geq 1$ **then**
          $deviatedValue =$
            $\min_{(p,r) \in \mathcal{U}_s} q_t(s,d-1,a,p,r)$
          **if** $actionValue > deviatedValue$ **then**
            $actionValue = deviatedValue$
          **end if**
        **end if**
        **if** $bestActionValue < actionValue$ **then**
          $bestActionValue = actionValue$
          $bestAction = a$
        **end if**
      **end for**
      $V_t(s,d) = bestActionValue$
      $\pi_t(s,d) = bestAction$
    **end for**
  **end for**
**end for**
**return** $\pi_t(s,d) \quad \forall t \in [1:T], s \in \mathcal{S}, d \in [0:D]$

In Setup A the total number of stages where parameters deviate is bounded. In contrast, in Setup B, similarly to future reward, future parameter deviation is discounted. Setup A is easy to compute as we can apply the same technique to augment the state-space with the remaining budget of "Nature". Setup B further requires discretization of the remaining budget. Both setups' optimal value functions admit implicit equations similar to the Bellman Equations, as shown in Appendix F in the supplementary material, and thus are solvable in the same methods as classic infinite-horizon MDPs.

## 5. Continuous Deviations

This section is motivated from the following setup: in each stage, the real parameter is likely to *slightly* deviate from the nominal one, whereas large deviation is rare. Notice that the previous approach that bounds the number that parameter deviates does not apply to this setup. Hence, we extend previous results to the continuous deviation case, i.e., Nature is allowed to perform "fractional" deviations.

More specifically, suppose the nominal parameters are $p^0, r^0$ and the uncertainty set is $\mathcal{U}$, and that $\mathcal{U}_s$ is star-shaped[2] in respect to $(p_s^0, r_s^0)$ for each $s \in \mathcal{S}$. If the realized parameter is $(p, r) \in \mathcal{U}$, then Nature is considered to have consumed a budget of deviation $b(p, r, p^0, r^0, \mathcal{U})$ such that

$$b(p, r, p^0, r^0, \mathcal{U}) \triangleq \min\{\alpha \geq 0 \,|\,$$
$$\exists (\delta_p, \delta_r) \in \Delta\mathcal{U} : (p, r) = (p^0 + \alpha\delta_p, r^0 + \alpha\delta_r)\},$$

where $\Delta\mathcal{U} \triangleq \{(\delta_p, \delta_r) | (p^0 + \delta_p, r^0 + \delta_r) \in \mathcal{U}\}$.

Since $(p, r) \in \mathcal{U}$, we get $b(p, r, p^0, r^0, \mathcal{U}) \leq 1$. Roughly speaking, a "small" deviation is considered "less expensive" compared to a large one. Thus, we can formulate the continuous extension as follows, with $\gamma \leq \beta \leq 1$, and $\gamma < 1$.[3]

**Continuous Deviations:**

$$\max_{a_1 \in \mathcal{A}} \min_{(p_1, r_1) \in \mathcal{U}_{s_1}} \max_{a_2 \in \mathcal{A}} \min_{(p_2, r_2) \in \mathcal{U}_{s_2}} \cdots \quad (6)$$

$$\max_{a_t \in \mathcal{A}} \min_{(p_t, r_t) \in \mathcal{U}_{s_t}} \cdots \mathbb{E}\left\{\sum_{t=1}^{\infty} \gamma^{t-1} r_t(s_t, a_t)\right\};$$

$$\text{Subject to:} \quad \sum_{t=1}^{\infty} \beta^{t-1} b(p_t, r_t, p_t^0, r_t^0, \mathcal{U}_{s_t}) \leq D.$$

Similarly to Problems 4 and 5 from the previous section, Problem 6 is solvable (approximately) using a discretization scheme. The statement and proof are given in Appendix G in the supplementary material.

We next provide a probabilistic guarantee of the continuous model, for the finite horizon case. The proof is similar to the previous probabilistic guarantees and is given in Appendix H in the supplementary material.

**Theorem 6.** *Suppose the amount of parameter deviation, $b(p_t, r_t, p_t^0, r_t^0, \mathcal{U}_{s_t})$ is independent with a mean $\alpha_t$. Then with probability at least $1 - \delta$ we have the that total amount of deviation, $\sum_{t=1}^{T} b(p_t, r_t, p_t^0, r_t^0, \mathcal{U}_{s_t})$, is upper bounded by $\sum_{t=1}^{T} \alpha_t + \frac{1}{3}\log(1/\delta)\left(1 + \sqrt{1 + 18\frac{\sum_{t=1}^{T}\alpha_t}{\log(1/\delta)}}\right)$.*

## 6. Simulations

We present simulation results of the proposed method under a setup closely resembles our motivating example, i.e., parameters randomly deviate from their nominal values with a small probability. We use a scenario based on the Single-Product Stochastic Inventory Control problem described in Puterman (1994). Here the states represent the number of items in the inventory, with maximal capacity of $MAXSTOCK$ items. Every day an order is made for new items at the cost of $STOREPRICE$ each. A price is paid for holding the items in the inventory (both old and new), which is a function of the number of items $HOLDINGCOST(n)$. A Poisson-distributed number of customers, with expected value of $NUMCUSTOMERS$ place orders for the item and pay $CUSTOMERPRICE$ for each product sold. If the demand cannot be met, the store is "fined" total of $PENALTY(n)$ where $n$ is the number of customers whose demand cannot be met. The simulation is run for $T$ days, and any unused stock at time $T + 1$ is wasted.

The deviation we added to this scenario is *Rush*: the maximal number of customers, $MAXSTOCK$, arrive simultaneously on the same day. Thus the uncertainty set $\mathcal{U}_{s_t}$ consists of two points: regular Poisson arrival and *Rush*. Note that a *Rush* affects both the transition probabilities and the rewards.

We simulated the above scenario for different values of initially assumed number of deviations $d_0$[4], and applied random occurrences of *Rush* with probability $p_{rush}$ for each day independently. We also added the expected value of the optimal MDP policy *aware* of the occurrence of *Rush* for comparison. The results [5] are given in Figure 3. It can be seen that with a suitable choice of $d_0$ such as expected number of deviation, the performance of LDST's policy is comparable to the performance of the optimal policy for each scenario, while both nominal ($d_0 = 0$) and conservative, uncoupled uncertainty ($d_0 = T$) polices give worse performance as expected.

## 7. Conclusion

We proposed a new robust MDP framework, termed "Lightning Does not Strike Twice," to model the case that uncertain parameters among different states are coupled by sharing a common pool of "budget" of deviation. This leads to a tractable formulation that provides a flexible tradeoff between risk and value, which can be adjusted by tuning the "budget."

---

[2] A set $\mathcal{U}$ is star-shaped w.r.t. $u_0$ if for any $u \in \mathcal{U}$, the line segment between $u$ and $u_0$ also belongs to $\mathcal{U}$.

[3] Notice that both the finite horizon case, and the two infinite horizon cases investigated, can be formulated in this way.

[4] Note that for $d_0 = 0$ we get the nominal policy, and for $d_0 = T$ we get the conservative, uncoupled uncertainty policy.

[5] The parameters chosen for simulation are $T = 100$, $MAXSTOCK = 20$, $STOREPRICE = 5$, $CUSTOMERPRICE = 50$, $NUMCUSTOMERS = 10$, $HOLDINGCOST(n) = 2n^2$, $PENALTY(n) = 7n^2$, initially the stock is empty.

Lightning Does Not Strike Twice: Robust MDPs with Coupled Uncertainty

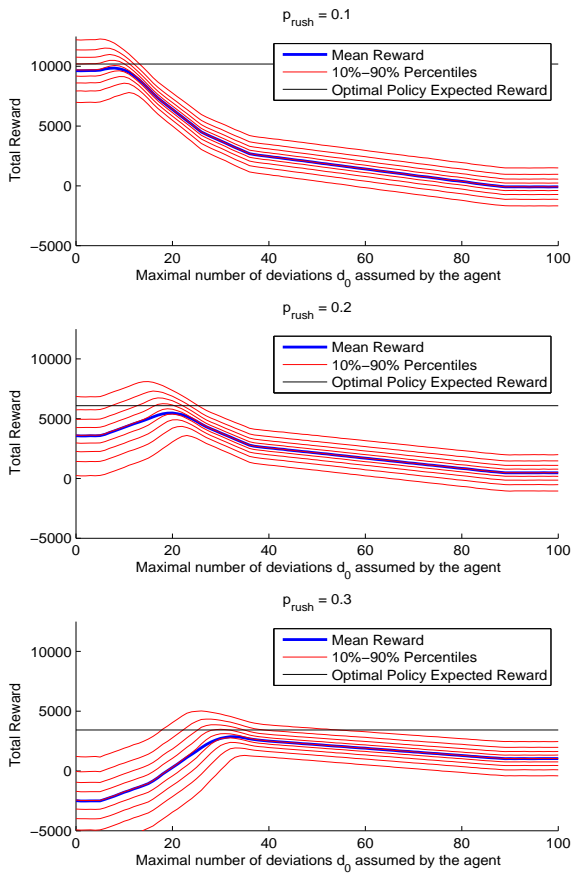

*Figure 3.* Simulation results for three values of $p_{rush}$. The number of simulated trajectories is $10^5$.

One obstacle that prevents robust MDPs from being widely applied is that the solution tends to be conservative – an outcome of the traditional formulation that uncertainties are uncoupled. On the other hand, coupled uncertainty in MDPs is in general computationally difficult. Therefore, it is important to identify sub-classes of coupled uncertainty that are flexible enough to overcome conservativeness, yet remains computationally tractable. This paper is the first of such attempts, which we hope will facilitate the applicability of robust MDPs.

## Acknowledgments

S. Mannor has received funding from the Israel Science Foundation (contract 890015) and the European Unions Seventh Framework Programme (FP7/2007-2013) under grant agreement no. 249254. H. Xu is partially supported by the National University of Singapore under startup grant R-265-000-384-133.